\documentclass[sigconf, nonacm]{acmart}
\usepackage[commandnameprefix=always]{changes}
\usepackage{xcolor}
\usepackage{tcolorbox}
\usepackage{alltt}
\usepackage{array, makecell}
\usepackage[utf8]{inputenc} 
\usepackage[T1]{fontenc}    
\usepackage{hyperref}       
\usepackage{url}            
\usepackage{booktabs}       
\usepackage{amsfonts}       
\usepackage{nicefrac}       
\usepackage{microtype}      
\usepackage{float}
\usepackage{natbib}
\usepackage{doi}
\usepackage{xcolor}
\usepackage{lineno}
\usepackage{amsmath}
\usepackage{enumitem}
\usepackage{changes}
\usepackage{url}
\newcommand\myshade{85}
\colorlet{mycitecolor}{violet}
\colorlet{myurlcolor}{blue}

\hypersetup{
  citecolor  = mycitecolor!\myshade!black,
  urlcolor   = myurlcolor!\myshade!black,
  colorlinks = true,
}

\usepackage{color}

\usepackage{times}
\usepackage{latexsym}
\usepackage{appendix}

\usepackage[noabbrev,capitalize]{cleveref}
\crefname{lstlisting}{listing}{listings}

\newcommand{\codefont}{\fontfamily{lmtt}\selectfont}
\usepackage{soul}
\definecolor{aigold}{RGB}{244,210, 1} 
\definecolor{aigreen}{RGB}{245, 255, 249}

\definecolor{humanpurple}{RGB}{235, 222, 240} 

\definecolor{commentgray}{RGB}{86, 101, 115}

\definecolor{aired}{RGB}{255,180,181}

\usepackage{listings}
\usepackage{parcolumns}
\lstdefinestyle{datalogstyle}{
	basicstyle={\codefont\small},  
	xleftmargin={6pt},
        xrightmargin={6pt},
        breakindent=0pt,
	frame=tb,
	stepnumber=1,
	firstnumber=1,
	numberfirstline=true,
	tabsize=2,
	showtabs=false,
	showspaces=false,
	showstringspaces=false,
	extendedchars=true,
	breaklines=true,
	columns=fullflexible,
	keepspaces=true,
	escapeinside={@}{@},
	firstnumber=last,
	captionpos=b,
	commentstyle=\color{black!65},
	numberstyle=\tiny\color{black!65},
	stringstyle=\color{codepurple},
	breakatwhitespace=false, 
	keepspaces=true,                 
	numbersep=5pt,                  
	showspaces=false,                
	showstringspaces=false,
	showtabs=false,
	aboveskip={0.8\baselineskip},
	belowskip={0.2\baselineskip},
	backgroundcolor=\color{aigreen},
}
\usepackage{tabularx, booktabs}
\newcolumntype{C}{>{\centering\arraybackslash}X}
\newcolumntype{R}{>{\raggedleft\arraybackslash}X}
\newcolumntype{S}{>{\raggedleft\arraybackslash\hsize=.5\hsize}X}

\usepackage{arydshln}
\usepackage{booktabs}
\usepackage{multirow}

\usepackage{pifont}  %
\usepackage{makecell}
\newcommand{\cmark}{\textcolor{green}{\ding{51}}}%
\newcommand{\xmark}{\textcolor{red}{\ding{55}}}%

\usepackage[T1]{fontenc}

\usepackage[utf8]{inputenc}

\usepackage{microtype}

\usepackage{fdsymbol}
\usepackage{booktabs}
\usepackage{graphicx}
\usepackage{makecell}
\usepackage{multirow}
\usepackage{longtable}
\usepackage{CJKutf8}
\usepackage{float}
\usepackage{algpseudocode}
\usepackage[frozencache, cachedir=minted-cache]{minted}
\usepackage[normalem]{ulem}

\usepackage{bm}

\usepackage{tikz}
\usetikzlibrary{shapes.geometric}

\usepackage[linesnumbered,ruled,vlined]{algorithm2e}

\SetCommentSty{mycommfont}

\usepackage{multirow}
\usepackage{csquotes}
\usepackage{xcolor}
\usepackage{amsmath}

\definecolor{ForestGreen}{RGB}{34,139,34}

\newcommand{\stitle}[1]{\vspace{2pt}\noindent\textbf{#1}}

\theoremstyle{plain}

\renewcommand\footnotetextcopyrightpermission[1]{} 

\begin{document}
\pagestyle{plain} 

\title{Demonstration of DB-GPT: Next Generation Data Interaction System Empowered by Large Language Models}

\author{%
  Siqiao Xue$^{\diamondsuit}$, Danrui Qi$^{\clubsuit}$, Caigao Jiang$^{\diamondsuit}$, Wenhui Shi$^{\diamondsuit}$, Fangyin Cheng$^{\varheartsuit}$, Keting Chen$^{\diamondsuit}$, \\ Hongjun Yang$^{\diamondsuit}$, 
  Zhiping Zhang$^{\heartsuit}$, Jianshan He$^{\diamondsuit}$, Hongyang Zhang$^{\vardiamondsuit}$, Ganglin Wei$^{\diamondsuit}$, \\ Wang Zhao,
  Fan Zhou$^{\diamondsuit}$, Hong Yi, Shaodong Liu$^{\spadesuit}$, Hongjun Yang$^{\diamondsuit}$, Faqiang Chen$^{\diamondsuit,*}$
  }
  \affiliation{%
  \institution{$^{\diamondsuit}$Ant Group, $^{\heartsuit}$Alibaba Group, $^{\varheartsuit}$ JD Group, $^{\spadesuit}$Meituan, \\
  $^{\vardiamondsuit}$Southwestern University of Finance and Economics, China, \\
  $^{\clubsuit}$Simon Fraser University, Canada}
 }

\begin{abstract}
The recent breakthroughs in large language models (LLMs) are positioned to transition many areas of software. The technologies of interacting with data particularly have an important entanglement with LLMs as efficient and intuitive data interactions are paramount. In this paper, we present DB-GPT, a revolutionary and product-ready Python library that integrates LLMs into traditional data interaction tasks to enhance user experience and accessibility. 
DB-GPT is designed to understand data interaction tasks described by natural language and provide context-aware responses powered by LLMs, making it an indispensable tool for users ranging from novice to expert.
Its system design supports deployment across local, distributed, and cloud environments. Beyond handling basic data interaction tasks like Text-to-SQL with LLMs, it can handle complex tasks like generative data analysis through a Multi-Agents framework and the Agentic Workflow Expression Language (AWEL). The Service-oriented Multi-model Management Framework (SMMF) ensures data privacy and security, enabling users to employ DB-GPT with private LLMs. Additionally, DB-GPT offers a series of product-ready features designed to enable users to integrate DB-GPT within their product environments easily. The code of DB-GPT is available at
Github\footnote{https://github.com/eosphoros-ai/DB-GPT}
which already has \textbf{over 10.7k stars}. Please install DB-GPT for your own usage with the instructions~\footnote{https://github.com/eosphoros-ai/DB-GPT\#install}
and watch a 5-minute introduction video on Youtube~\footnote{https://youtu.be/n\_8RI1ENyl4} to further investigate DB-GPT. 

\end{abstract}

\maketitle

\vspace{-.5em}
\section{Introduction}

Large language models (LLMs) such as ChatGPT and GPT-4 have showcased their remarkable capabilities in engaging in human-like communication and understanding complex queries, bringing a trend of incorporating LLMs in various fields.
Data interaction, which aims to let users engage with and understand their data, enabling the retrieval, analysis, manipulation, and visualization of data to derive insights or make decisions. In the realm of interacting with data, LLMs pave the way for natural language interfaces, enabling users to express their data interaction tasks through natural language and leading to more natural and intuitive data interactions.

Nonetheless, how to enhance the data interaction tasks with LLMs to provide users reliable understanding and insights to their data still remains an open question. One straightforward approach is to directly provide commonly used LLMs, such as GPT-4, with instructions on how to interact via few-shot prompting or in-context learning. Moreover, to further facilitate the intelligent interactions with data, many works~\citep{langchain,Liu_LlamaIndex_2022,autogpt} have incorporated the LLM-powered automated reasoning and decision process (a.k.a., multi-agents frameworks) into the data interaction process. However, these multi-agents frameworks are usually task-specific instead of task-agnostic, limiting their usage to a broad range of tasks. Meanwhile, the interaction with data includes a variety of tasks in practice. For example, it includes the Text-to-SQL / SQL-to-Text tasks, the generation of data analytics, the generation of enterprise report analysis and business insights, etc. It is necessary for users to arrange the workflow of multi-agents according to their own needs. The existing effort~\cite{langchain} does not consider abundant data interaction needs. Finally, though being important, the privacy-sensitive setup for LLM-empowered data interaction is under-investigated. The previous efforts~\citep{Martinez_Toro_PrivateGPT_2023,h2ogpt2023} are not designed for data interaction tasks.

To overcome these limitations, our key idea is to propose an open-sourced Python library \textsc{\textit{DB-GPT}} supporting data interaction by using multi-agents with flexible arrangement and privacy-sensitive setup. 
This idea, however, introduces three main challenges, the first challenge (\textbf{\textit{C1}}) is the design of multi-agents framework for supporting database interaction. 
The second challenge (\textbf{\textit{C2}}) is the declarative expression supporting arrange multi-agents flexibly. The third challenge (\textbf{\textit{C3}}) focuses on the design of private LLM-empowered data interaction.   

\textbf{\textit{To solve C1}}, we propose the Multi-Agents framework in \textsc{\textit{DB-GPT}} which automates the database interaction tasks. Once users have entered their final goals, the Multi-Agents framework can free their hands, autonomously generate the planning of tasks and execute particular tasks. 
\textbf{\textit{To solve C2}}, we proposes a declarative language called \textit{Agentic Workflow Expression Language (AWEL)} in \textsc{\textit{DB-GPT}}. With AWEL, users can implement their execution plan for multi-agents with simple expression (i.e. few lines of code). Furthermore, to make users more code-free, \textsc{\textit{DB-GPT}} also provides an interface for users constructing their Agentic Workflow with only drag and drop. 
\textbf{\textit{To solve C3}}, we propose \textit{Service-oriented Multi-model Management Framework (SMMF)} in \textsc{\textit{DB-GPT}} to support users to run \textsc{\textit{DB-GPT}} with their private LLMs in their own execution environment. All the interactions among users, LLMs and data are performed locally, which definitely promises users' privacy.

\begin{table*}[t]
\small
\caption{Comparasion between DB-GPT and other tools.}
\vspace{-1em}
\begin{tabular}{c|lccccc}
\hline
\textbf{}                                                                                            & \multicolumn{1}{c}{\textbf{}}      & \textbf{LangChain~\citep{langchain}}   & \textbf{LlamaIndex~\citep{Liu_LlamaIndex_2022}}  & \textbf{PrivateGPT~\citep{Martinez_Toro_PrivateGPT_2023}}  & \textbf{ChatDB~\citep{hu2023chatdb}}      & \textbf{DB-GPT}      \\ \hline
\multirow{5}{*}{\textbf{System Components}}       & Multi-Agents Framework & \cmark  & \cmark    & \xmark     & \xmark    & \cmark \\
    & Multi-LLMs Support & \cmark & \cmark & \xmark & \cmark & \cmark   \\
    & RAG from Multiple Data Sources & \cmark  &  \cmark & \xmark & \xmark    & \cmark   \\
    & Agent Workflow Expression Language & \xmark & \xmark & \xmark & \xmark & \cmark  \\
    & Fine-tuned Text-to-SQL Model & \xmark  & \cmark       & \xmark  & \xmark & \cmark  \\ \hline
\multirow{4}{*}{\textbf{\begin{tabular}[c]{@{}c@{}}Data Interaction\\ Functionalities\end{tabular}}} 

& Text-to-SQL / SQL-to-Text         & \cmark       & \cmark       & \xmark  & \cmark & \cmark  \\
& Chat2DB / Chat2Data / Chat2Excel   & \cmark       & \cmark       & \xmark  & \cmark & \cmark \\
& Data Privacy and Security   & \xmark       & \xmark       & \cmark  & \xmark & \cmark                      \\
& Multilingual Interactions   & \xmark       & \xmark       & \xmark  & \cmark & \cmark                      \\
& Generative Data Analysis  & \xmark       & \xmark       & \xmark  & \xmark & \cmark  \\ \hline
\end{tabular}
\label{tab:comparison}
\end{table*}

Additionally, the \textsc{\textit{DB-GPT}} community extends its support beyond basic functionalities, offering a suite of product-ready features designed to enhance data interaction capabilities. These include advanced knowledge extraction from diverse data sources for more accurate answers to users' queries, specialized fine-tuning of Text-to-SQL Large Language Models (LLMs) to facilitate seamless database queries, and a user-friendly front-end interface for more convenient interaction.  Furthermore, \textsc{\textit{DB-GPT}} supports multilingual functionality, accommodating both English and Chinese, thereby broadening its applicability and ease of use across different linguistic contexts. With these comprehensive, product-ready considerations, \textsc{\textit{DB-GPT}} is equipped to handle intricate data interaction tasks, such as generative data analysis, enabling users to seamlessly integrate and leverage its powerful functionalities within their product environments. This holistic approach ensures that \textsc{\textit{DB-GPT}} is not just a library, but a complete solution for developers and businesses aiming to harness the full potential of AI in the process of interacting with data. Table~\ref{tab:comparison} shows the comparison between \textsc{\textit{DB-GPT}} and other popular tools from two main perspectives: system components and data interaction functionalities, showing the superiority of \textsc{\textit{DB-GPT}}.



\begin{figure}[t]
\begin{center}
\centerline{\includegraphics[width=1.2\linewidth]{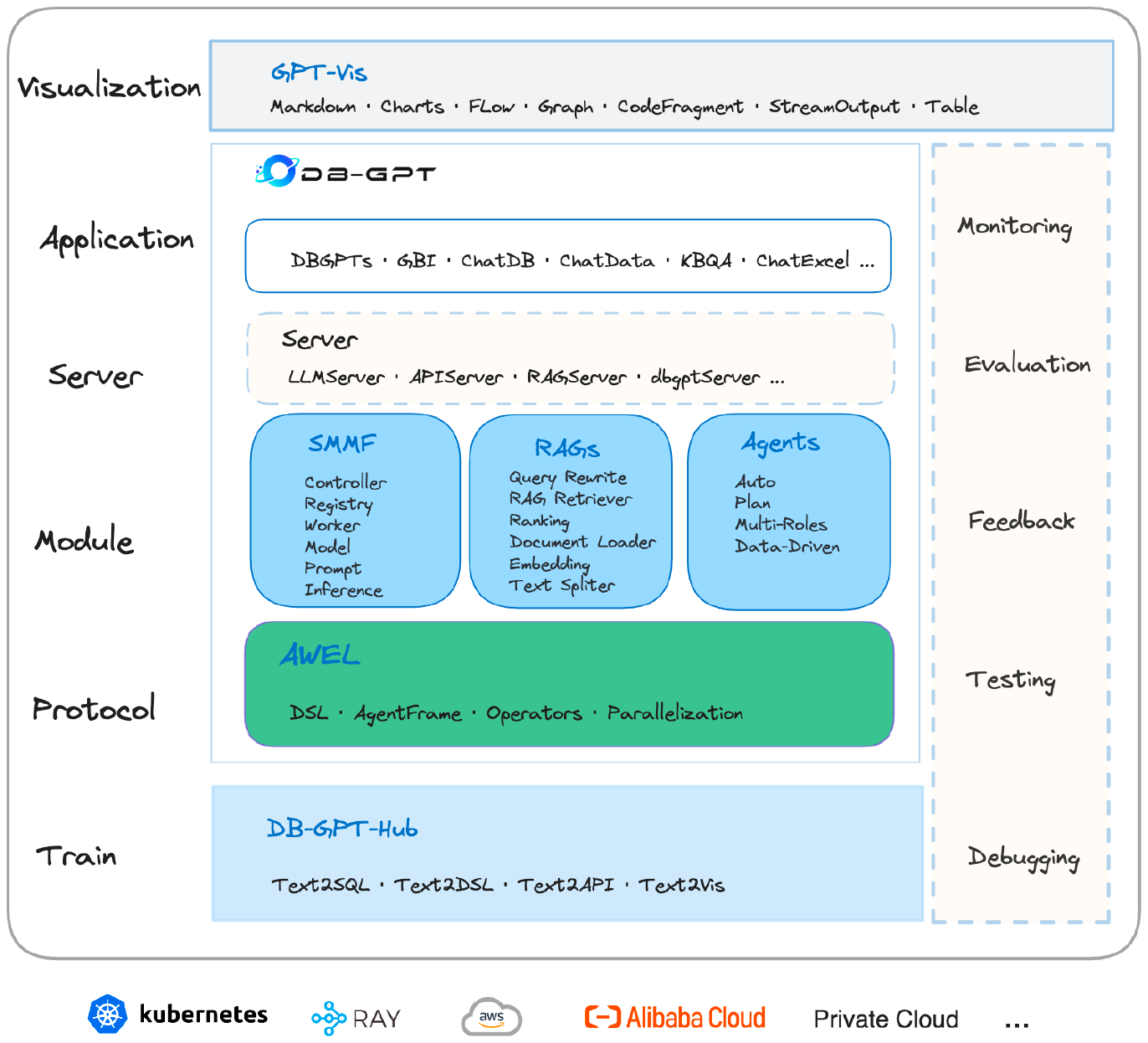}}
\vspace{-1em}
\caption{System Design of DB-GPT}
\label{fig:dbgpt}
\end{center}
\vspace{-1em}
\end{figure}

To summarize, we make the following contributions: 
 1) we propose \textsc{\textit{DB-GPT}}, an open-sourced and product-ready library supporting an end-to-end interaction with data. 
 2) we propose Multi-Agents Framework in \textsc{\textit{DB-GPT}} for solving complex data interaction tasks like generative data analysis. 
 3) we propose \textit{Agentic Workflow Expression Language (AWEL)} to enhance the practicability and flexibility of Multi-Agents in \textsc{\textit{DB-GPT}}.  
 4) we propose \textit{Service-oriented Multi-model Management Framework (SMMF)} to promise the users' privacy from the model perspective in \textsc{\textit{DB-GPT}}.
 5) we deploy \textsc{\textit{DB-GPT}} as an application with user-friendly interface and demonstrate its utility. We also open-sourced the implementation of \textsc{\textit{DB-GPT}} on \href{https://github.com/eosphoros-ai/DB-GPT}{\textcolor{blue}{Github}}, which already has \textbf{over 10.7k stars}.

\vspace{-.3em}
\section{System Design}

The overall system design of \textsc{\textit{DB-GPT}} is depicted in Figure~\ref{fig:dbgpt}. \textsc{\textit{DB-GPT}} includes four layers, i.e. the protocol layer, the module layer, the server layer and the application layer. In this section, we delineate the design of each phase with a top-down manner. There are also other layers making \textsc{\textit{DB-GPT}} product-ready. We also introduce the design of these layers in this section.

\vspace{-.3em}
\subsection{The Application Layer}
The application layer encompasses the array of data interaction functionalities supported by DB-GPT. These include, but are not limited to, Text-to-SQL/SQL-to-Text, chat-to-database interactions (chat2db), chat-to-data queries (chat2data), chat-to-Excel operations (chat2excel), chat-to-visualization commands (chat2visualization), generative data analysis, and question answering based on knowledge bases. These functionalities include the majority of foundational tasks associated with data interaction, illustrating the comprehensive capabilities of the DB-GPT framework. 

\vspace{-.3em}
\subsection{The Server Layer}
The server layer in \textsc{\textit{DB-GPT}} is an optional component that manages external inputs, such as HTTP requests, by integrating them with domain knowledge to guide lower-tier layers, i.e. the Module Layer. This layer's optional status allows for direct communication between the application layer and the module layer in simple scenarios. In contexts that necessitate external inputs, the server layer acts as a supplementary intermediary, underscoring its utility in supporting a wider range of applications.

\vspace{-.3em}
\subsection{The Module Layer}
The module layer of \textsc{\textit{DB-GPT}} is composed by \textit{Service-oriented Multi-model Management Framework (SMMF), Retrieval-Augmented Generation (RAG) from Multiple Data Source} and \textit{Multi-Agents Framework}. The three parts of the module layer are most important to support users' interaction with their data, which is shown in the Application Layer.

\stitle{Service-oriented Multi-model Management Framework (SMMF).}
The Service-oriented Multi-model Framework (SMMF) in the context of DB-GPT aims at facilitating model adaptation, enhancing deployment efficiency, and optimizing performance. SMMF offers a streamlined platform for the deployment and inference of Multi-Large Language Models (Multi-LLMs), enabling local execution of users' own LLMs to ensure data privacy and security.

\begin{figure*}[ht]
\begin{center}
\centerline{\includegraphics[width=0.85\linewidth]{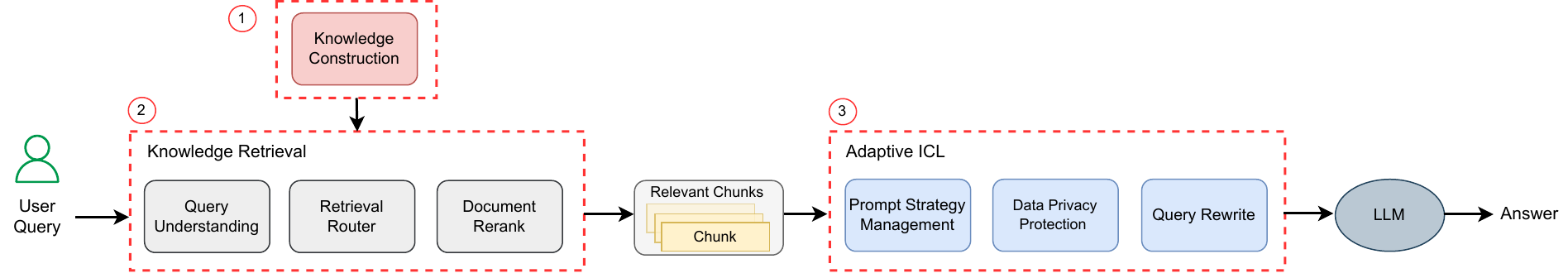}}
\caption{The RAG architecture in DB-GPT}
\label{fig:rag_arch}
\end{center}
\vspace{-1em}
\end{figure*}

SMMF is underpinned by two core components: the model inference layer and the model deployment layer. The inference layer supports various LLM inference frameworks, enhancing the framework's flexibility. The deployment layer connects inference mechanisms with model serving capabilities, incorporating an API server and a model handler for robust functionality. At its core, the model controller manages metadata, integrating the deployment process, while the model worker establishes connectivity with inference and infrastructure, ensuring efficient model operation. Through SMMF, DB-GPT provides an efficient approach to deploying machine learning models in a cloud environment, highlighting the framework's potential in improving adaptability, performance, and data security in MaaS applications.

\stitle{Retrieval-Augmented Generation (RAG) from Multiple Data Source.} While LLMs are usually trained on enormous bodies of open sourced or other parties' proprietary data, RAG~\cite{Lewis2020RetrievalAugmentedGF} is a technique for augmenting LLMs' knowledge with additional and often private data. Shown in Figure~\ref{fig:rag_arch}, our RAG pipeline consists of three stages: knowledge construction, knowledge retrieval and adaptive In-Contextual Learning (ICL)~\cite{Dong2022ASO} strategies.

For knowledge construction, DB-GPT constructs a knowledge base according to multiple data sources provided by users. Contents in each data source are segmented into paragraphs, with each paragraph encoded into a multidimensional vector using a neural encoder. Notably, DB-GPT enhances traditional vector-based knowledge representation by integrating inverted index and graph index methods, facilitating precise context-relevant data retrieval.
For knowledge retrieval, upon receiving a query $x$, it is transformed into a query vector $q$. DB-GPT then identifies the top-$k$ paragraphs within the knowledge base that are most relevant to $q$. DB-GPT employs diverse retrieval strategies for prioritizing relevant documents, including ordering based on the cosine similarity of their embedded vectors, as well as categorization according to keyword similarity.
In the adaptive iterative contextualization phase, DB-GPT employs Interactive Contextual Learning (ICL) for generating responses. ICL enhances DB-GPT's response by integrating knowledge retrieval results during LLMs' inference. It incorporates them into a predefined prompt template to get response from LLM. The efficacy of ICL depends on specific configurations such as prompt templates. Our DB-GPT system provides various strategies for prompt formulation and incorporates privacy measures to protect private information.  Due to the page limit, please see~\citep{xue2023dbgpt}  for the full details.

\stitle{Multi-Agents Framework.} Inspired by 
\href{https://github.com/geekan/MetaGPT}{\textcolor{blue}{MetaGPT}} and \href{https://microsoft.github.io/autogen/}{\textcolor{blue}{AutoGen}}, when dealing with challenging data interaction tasks such as generative data analysis, DB-GPT proposes its own Multi-Agent framework. The proposed framework leverages the specialized capabilities and communicative interactions of multiple agents to effectively address multifaceted challenges. For example, consider the task of constructing detailed sales reports from at least three distinct dimensions. The Multi-Agent framework initiates this process by deploying a planning agent to devise a comprehensive strategy, which includes the creation of: 1) a donut chart for the analysis of total sales by product category, 2) a bar chart for examining sales data from the perspective of user demographics, and 3) an area chart for evaluating monthly sales trends. Subsequent to the planning phase, dedicated chart-generating agents are tasked with the production of these visual representations, which are then aggregated by the planner and presented to users.

Compared to MetaGPT and AutoGen, DB-GPT's Multi-Agent framework archives the entire communication history among its agents within a local storage system, thereby significantly enhancing the reliability of the generated content of agents. Furthermore, in contrast to the LlamaIndex framework, which prescribes a set of constrained behaviours tailored to specific use cases, DB-GPT's framework offers flexibility which allows users to custom-define agents tailored to their specific data interaction tasks, thus affording a broader applicability across various domains.

\vspace{-.3em}
\subsection{The Protocol Layer}
The protocol layer in \textsc{\textit{DB-GPT}} mainly includes \textit{Agentic Workflow Expression Language (AWEL)}, 
which adopts the big data processing concepts of \href{https://airflow.apache.org/}{\textcolor{blue}{Apache Airflow}}. By leveraging Directed Acyclic Graphs (DAGs), AWEL orchestrates workflows, aligning with Apache Airflow's mission to efficiently define, schedule, and oversee complex data pipelines and workflows.

In Apache Airflow, the core components of these workflows are operators, where each operator represents a discrete task or operation capable of executing defined actions. Reflecting this approach, \textsc{\textit{DB-GPT}}'s AWEL models each agent as a distinct operator, thus enabling users to intricately design their agent-based workflows. This is achieved by interconnecting multiple agents to construct a DAG. Such a design grants users the flexibility to manipulate the flow of information between agents. Consequently, users can seamlessly integrate their comprehension of specific data interaction tasks with the actionable insights generated by LLM-based agents. 

Employing AWEL within \textsc{\textit{DB-GPT}} empowers it to support a variety of tasks including stream processing, batch processing, and asynchronous operations. This capability significantly bolsters \textsc{\textit{DB-GPT}}'s effectiveness and applicability in navigating the complexities of real-world production environments.

\begin{figure*}[t]
\begin{center}
\centerline{\includegraphics[width=\linewidth,scale=0.5]{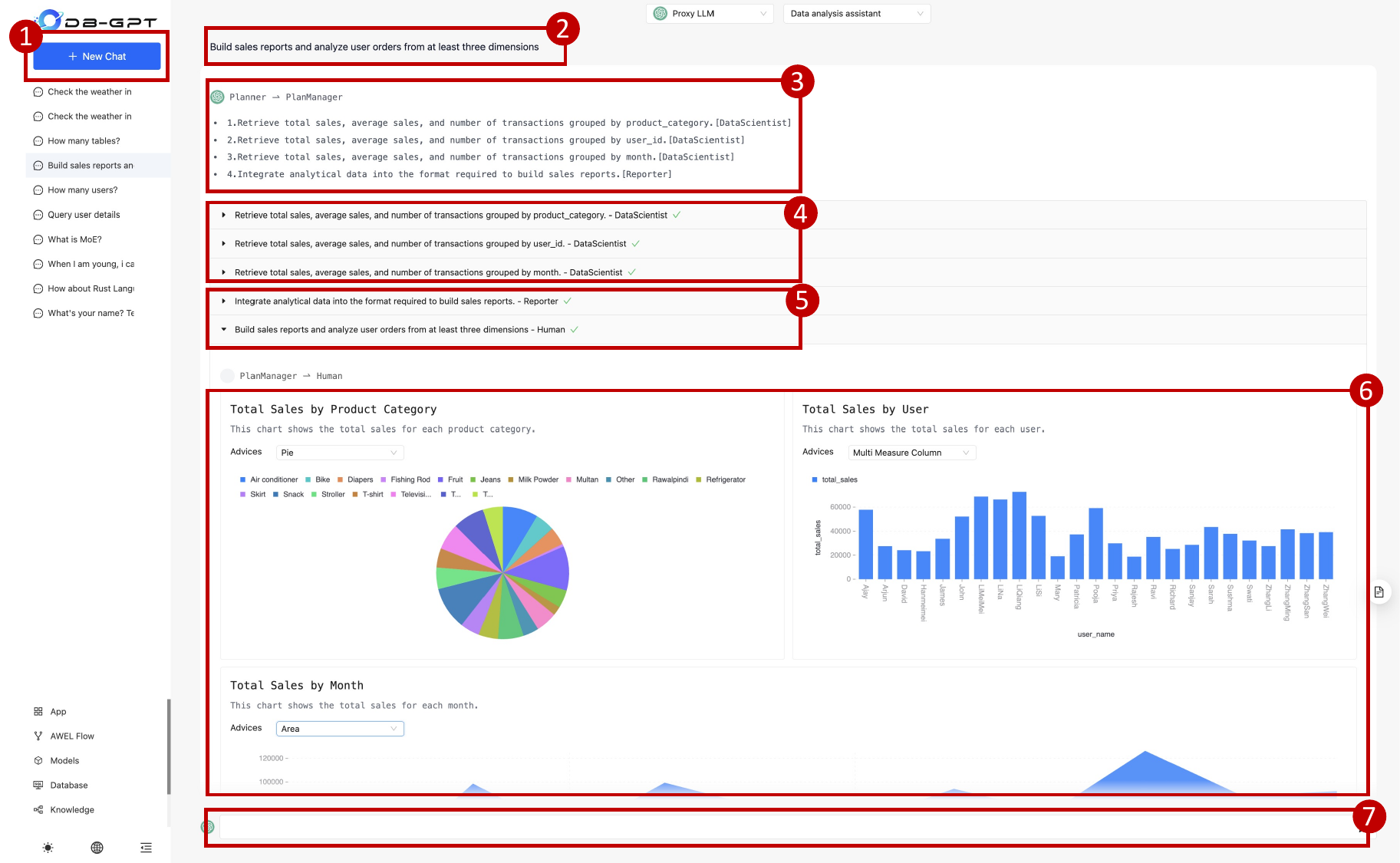}}
\vspace{-.8em}
\caption{Demonstration of DB-GPT}
\label{fig:demo}
\end{center}
\vspace{-1em}
\end{figure*}

\vspace{-.3em}
\subsection{Other Layers}
\stitle{Visualization Layer.}
The visualization layer aims to display the answers returned by DB-GPT to the users with elegance. For scenarios involving purely textual question-and-answer formats, this layer exhibits the textual outputs generated by DB-GPT. When the tasks necessitate the generation of charts, DB-GPT renders these charts within its front-end, facilitating user interaction with the displayed charts. The significance of possessing a sophisticated visualization layer within DB-GPT cannot be overstated, as practical applications demand that users engage in multiple interactions with their data to successfully complete their data interaction tasks. Superior visualization capabilities significantly expedite users' comprehension of their data, thereby enhancing the overall effectiveness and user experience of the system.

\stitle{Text-to-SQL Fine-Tuning.} Although LLMs,.e.g., CodeX and ChatGPT, have shown successful results for Text-to-SQL, they still have a gap with the fine-tuned alternatives in specific application scenarios. Consequently, tailoring LLMs to domain-specific Text-to-SQL datasets emerges as a crucial step towards enhancing their comprehension of prompts and facilitating superior outcomes.
Within DB-GPT, we have introduced a component, termed DB-GPT-Hub, which serves to encapsulate the Text-to-SQL fine-tuning process. This module enables users to refine their Text-to-SQL LLMs using publicly available LLMs hosted on Huggingface in conjunction with their own Text-to-SQL data pairs. Moreover, our SMMF framework accords users the flexibility to employ their fine-tuned LLMs in a localized manner, which underscores our commitment to data privacy and security.

\stitle{Execution Environments.} DB-GPT is capable of operating within distributed environments through the employment of the distributed framework Ray, as well as within cloud ecosystems such as AWS Cloud, Alibaba Cloud, and private cloud configurations maintained by users. This operational flexibility underscores DB-GPT's adeptness at accommodating a variety of data storage contexts. Furthermore, it exemplifies the wide applicability of DB-GPT across diverse operational requirements and environments.

\vspace{-.3em}
\section{Demonstration}

The demonstration setup includes a table need to be standardized and a laptop. The laptop must connect to the Internet for visitors can use DB-GPT smoothly with OpenAI's GPT service. Visitors can also choose local models such as Qwen and GLM. If the conference Internet fails, a mobile hotspot (established via cell phone) can also be used for running DB-GPT.

Figure~\ref{fig:demo} illustrates the capability of DB-GPT to perform generative data analysis. When users are faced with a data interaction task, they initiate the process by starting a new chat session (area \ding{172}) and entering a command such as "Build sales reports and analyze user orders from at least three distinct dimensions" (area \ding{173}). DB-GPT undertakes this task utilizing its Multi-Agent framework, which begins with invoking a planner to generate a four-step strategy tailored to the task (area \ding{174}). Then, three specialized agents, designated for the creation of data analytics charts, proceed to generate sales reports (area \ding{175}). These report takes into account various dimensions, including product category, user name and month. Another agent, dedicated to aggregating these charts, collects, organizes, and presents them on the front-end interface (area \ding{176}). The interface allows users to interact with the displayed charts, offering the flexibility to alter chart types according to their preferences (area \ding{177}). If users need further data interaction tasks to be performed, they can continue to engage with their data through natural language inputs (area \ding{178}).


\vspace{-.3em}
\section{Conclusion}
In this paper, we proposed DB-GPT, a revolutionary and product-ready Python library that understands data interaction tasks described by natural language and provides responses powered by LLMs. With the four-layer system design, DB-GPT can handle complex data interaction tasks with privacy consideration. In the future, DB-GPT will adapt more data interaction needs with its code-free agentic workflow.


Future research directions include: 1) introducing powerful agents providing more powerful abilities, such as time series predictions~\citep{jin2023large,xue2021graphpp,xue2022hypro,xue2023easytpp,shi2023language} based on historical data and predictive decision abilities~\citep{xue_meta_2022,qu-2022-rltpp} and automatic data preparation~\citep{Auto-FP, FeatAug, CleanAgent}; 2) the integration of more model training techniques. In addition to pre-training, the community is also interested in continual learning techniques for language models, such as continual pre-training~\citep{jiang2023anytime}, prompt learning~\citep{wang2022learning,xue2023prompttpp}.



\bibliographystyle{ACM-Reference-Format}
\bibliography{reference}

\end{document}